\documentclass{article}

\usepackage[preprint,nonatbib]{neurips_2025}
\usepackage[numbers]{natbib}
\usepackage{graphicx}
\usepackage{multirow}
\usepackage{amsthm}
\usepackage{wrapfig}
\theoremstyle{definition}

\usepackage[utf8]{inputenc} 
\usepackage[T1]{fontenc}    
\usepackage{hyperref}       
\usepackage{url}            
\usepackage{booktabs}       
\usepackage{amsfonts}       
\usepackage{amsmath}
\usepackage{nicefrac}       
\usepackage{microtype}      
\usepackage{xcolor}         
\usepackage{comment}
\usepackage{caption} 
\usepackage{subcaption}
\usepackage[nameinlink,capitalize]{cleveref}

\newif\ifincl
\incltrue

\title{DeepOSets: Non-Autoregressive In-Context Learning with Permutation-Invariance Inductive Bias}

\author{Shao-Ting Chiu\\
    Dept of Electrical \& Computer Engineering\\
  Texas A\&M University\\
  College Station, TX, USA
  \texttt{stchiu@tamu.edu}
  \And
Junyuan Hong\\
   Department of Electrical \& Computer Engineering\\
  University of Texas at Austin\\
  Austin, TX, USA
  \texttt{jyhong@utexas.edu} 
  \And
  Ulisses Braga-Neto\\
    Dept of Electrical \& Computer Engineering\\
  Texas A\&M University\\
  College Station, TX, USA
  \texttt{ulisses@tamu.edu}}

\def\v{\mathbf}
\def\m{\mathbf}
\def\bx{\mathbf{x}}
\def\sB{\mathcal{B}}
\def\sX{\mathcal{X}}
\def\sY{\mathcal{Y}}
\def\sZ{\mathcal{Z}}
\def\sH{\mathcal{H}}
\def\hnu{\hat{\nu}}
\def\BB{{\mathbb{B}}}
\def\RR{{\mathbb{R}}}

\theoremstyle{thmstyleone}%
\newtheorem{theorem}{Theorem}
\theoremstyle{thmstyletwo}%

\theoremstyle{thmstylethree}%

\begin{document}

\maketitle

\begin{abstract}
In-context learning (ICL) is the remarkable ability displayed by some machine learning models to learn from examples provided in a user prompt without any model parameter updates. ICL was first observed in the domain of large language models, and it has been widely assumed that it is a product of the attention mechanism in autoregressive transformers. In this paper, using stylized regression learning tasks, we demonstrate that  ICL can emerge in a non-autoregressive neural architecture with a hard-coded permutation-invariance inductive bias. This novel architecture, called DeepOSets, combines the set learning properties of the DeepSets architecture with the operator learning capabilities of Deep Operator Networks (DeepONets). 
We provide a representation theorem for permutation-invariant regression learning operators and prove that DeepOSets are universal approximators of this class of operators. We performed comprehensive numerical experiments to evaluate the capabilities of DeepOSets in learning linear, polynomial, and shallow neural network regression, under varying noise levels, dimensionalities, and sample sizes. 
In the high-dimensional regime, accuracy was enhanced by replacing the DeepSets layer with a Set Transformer.
Our results show that DeepOSets deliver accurate and fast results with an order of magnitude fewer parameters than a comparable transformer-based alternative.
\end{abstract}

\newcommand{\bc}{\begin{center}}
\newcommand{\bi}{\begin{itemize}}
\newcommand{\be}{\begin{enumerate}}
\newcommand{\beq}{\begin{equation}}
\newcommand{\bbm}{\begin{bmatrix}}
\newcommand{\bcs}{\begin{cases}}
\newcommand{\bal}{\begin{aligned}}
\newcommand{\ec}{\end{center}}
\newcommand{\ei}{\end{itemize}}
\newcommand{\ee}{\end{enumerate}}
\newcommand{\eeq}{\end{equation}}
\newcommand{\ebm}{\end{bmatrix}}
\newcommand{\ecs}{\end{cases}}
\newcommand{\eal}{\end{aligned}}
    
\def\v{\mathbf}
\def\m{\mathbf}
\def\bx{\mathbf{x}}
\def\sB{\mathcal{B}}
\def\sX{\mathcal{X}}
\def\sY{\mathcal{Y}}
\def\sZ{\mathcal{Z}}
\def\sH{\mathcal{H}}
\def\hnu{\hat{\nu}}
\def\BB{{\mathbb{B}}}
\def\RR{{\mathbb{R}}}
\def\prg{\vspace{2ex}\noindent}
\def\prgo{\vspace{1ex}\noindent}
\def\ul{\underline}
\def\vsp{\vspace{1ex}}
\def\eps{\varepsilon}
\def\sU{\mathcal{U}}
\def\sX{\mathcal{X}}
\def\sY{\mathcal{Y}}
\def\sZ{\mathcal{Z}}
\def\sH{\mathcal{H}}
\def\rt{\rightarrow}
\def\citep{\cite}

\def\vv{\boldsymbol}

\section{Introduction}

In-context learning (ICL) is a phenomenon in which a trained machine learning model learns from examples in a user prompt and predicts the response to a query without further training \citep{brown2020language}. ICL was first demonstrated in natural language processing using attention-based architectures~\citep{vaswani2017attention}. ICL is a form of {\em meta-learning}, or ``learning to learn''\citep{schmidhuber1987evolutionary,schmidhuber1993neural}, which is an idea that is present in several key areas of machine learning, including few-shot learning \citep{li2017meta}, multitask learning~\citep{crawshaw2020multi}, continuous learning~\citep{javed2019meta}, and foundation models~\citep{bommasani2021opportunities}. 
Stylized regression tasks have recently been extensively used to investigate ICL using attention-based architectures~\citep{akyurek2022learning,garg2023,zhang2024trained,bai2024transformers,xing2024benefits,liu2024can,von2023transformers}, as well as state-space (Mamba) \citep{grazzi2024mamba} and MLP-based \citep{tong2024mlps} approaches. With the exception of MLPs, all other mentioned architectures operate autoregressively: the prompt must be processed example by example in a sequential fashion. However, processing the prompt sequentially leads to low efficiency and loss of accuracy in the case of long prompts.
Moreover, many tasks, including regression, consume the training data (or at least
mini-batches) in parallel, not sequentially. 

In this paper, we demonstrate that ICL can be achieved by a non-autoregressive architecture with built-in permutation-invariance inductive bias. This new architecture, called DeepOSets, combines set learning, 
by means of a DeepSets layer \citep{zaheer2017deep} or, alternatively, a Set Transformer \citep{lee2019set}, 
and operator learning, via deep neural operators (DeepONets) \citep{li2020neural,lu2021learning,wang2021learning}.
The justification for a set learning approach is that in many tasks, including regression, examples in the prompt form a set: the order of examples is immaterial, and the number of examples can vary. Set learning exploits this by adding a permutation-invariance inductive bias to the neural architecture, which is advantageous, since the model does not need to learn that the problem is permutation-invariant. In addition, the set learning layer allows for prompts of any size, a capability that is absent in MLP-based architectures, and is achieved in autoregressive architectures by processing the prompt sequentially. By contrast, the set learning layer allows the prompt to be processed in parallel, in non-autoregressive fashion, resulting in more efficient computation and conferring robustness to long prompts. The justification for an operator learning approach is that the key distinction between ordinary regression and in-context regression is that ordinary regression simply learns a
function $f:\sX \rt \sY$ from the input space $\sX$ to the output space $\sY$, while in-context regression learns an {\em operator} $\Phi_n: {\cal D} \rt \sH$ between a data space ${\cal D}$, containing the training data in the prompt, to a hypothesis space $\sH$ of functions $f:\sX \rt \sY$. The term ``operator'' is used to indicate
that one of the spaces involved is a function space, namely, the
hypothesis space. 


The training data in the prompt is encoded by the set learning layer into a sufficiently large latent space, which is the input to the branch network of the DeepONet, while the query data point is the input of the trunk network of the DeepONet. The addition of the set learning encoder is the key modification that allows the DeepONet model to process a prompt of
variable size in fully parallel fashion. In addition, it confers
a permutation-invariant inductive bias that improves
generalizability, as mentioned previously. 

We prove that a continuous permutation-invariant ICL operator for regression tasks can be written as a composition of a permutation-invariant function encoder and an operator decoder with a sufficiently large latent space, which extends results for permutation-invariant functions in \cite{zaheer2017deep} and \cite{wagstaff2019limitations}. This provides further justification for the combination of a set-learning layer and a neural operator in our approach. We also prove that DeepOSets are universal approximators for continuous permutation-invariant ICL regression operators.

We present the results of a comprehensive set of experiments that test the performance of DeepOSets in ICL for stylized regression tasks used in \cite{akyurek2022learning,garg2023,zhang2024trained,bai2024transformers,xing2024benefits,liu2024can,von2023transformers}, including 
linear regression and shallow neural network regression with varying levels of noise in low ($d=1$) and high ($d=20$) dimensional spaces. In the high-dimensional experiment, we found that accuracy could be significantly boosted by employing a Set Transformer in place of the DeepSets layer. This has the downside of introducing quadratic complexity in the number of examples, in contrast to DeepSets, which has linear complexity, though this extra complexity can be alleviated by using inducing points, without sacrificing accuracy significantly. These experimental results indicated that DeepOSets produce accurate ICL results, which are indeed more accurate than a comparable autoregressive transformer-based method \cite{garg2023}, while being faster to train and employing an order of magnitude fewer weights. We also observed that DeepOSets scale better to long prompts, due to parallel processing of the prompt; the fact that autoregressive architectures scale poorly to long prompts had already been observed in \cite{grazzi2024mamba}. Finally, we present experimental results with polynomial regression that demonstrate that DeepOSets is capable of performing automatic model selection, where the order of the polynomial is identified in-context from the data in the prompt more accurately than using traditional cross-validation model selection.


Our main contributions in this paper can be summarized as follows:
\begin{itemize}
\item We demonstrate that ICL can emerge in non-autoregressive neural architectures with built-in permutation invariance inductive bias.

\item We introduce DeepOSets, a fully parallel, non-autoregressive neural architecture, which may be of independent interest for set learning and operator learning.

\item We prove that an ICL regression operator is continuous and permutation-invariant with respect to the examples in the prompt if and only if it can be written as a composition of a continuous permutation-invariant encoder followed by a continuous decoding operator. We also show that DeepOSets is a universal approximator for continuous permutation-invariant operators.

\item We show empirically that DeepOSets can perform ICL in stylized linear and shallow neural network regression tasks more efficiently and accurately than a comparable, well-known transformer-based approach. We also show empirically that DeepOSets can perform in-context model selection for choosing the order of polynomial regression more accurately than the traditional cross-validation approach.

\end{itemize}

\section{Methods}

\subsection{In-Context Learning in Regression Tasks}

As our goal is to demonstrate ICL in regression tasks, here we define formally ICL in this setting. Let $\sX \subseteq R^d$ and $\sY \subseteq R^p$ be input and
output spaces, and let $D_n = \{(\v{x}_1,y_1),\ldots,(\v{x}_n,y_n)\} \in {\cal D}_n$ be the sample data, where ${\cal D}_n = (\sX \times \sY)^n$ is the (finite-dimensional) data space. In regression, we can write $y_i = f(\v{x}_i) + \eps_i$, $i=1,\ldots,n$, where $f$ is the groundtruth function and $\eps$ is an additive noise variable. A {\em regression operator} is a mapping $\Phi_n: {\cal D}_n\rightarrow\sH$, where $\sH$ is a hypothesis function space containing all groundtruth functions of interest. Notice that $\Phi_n$ is not an ordinary regression function; it maps the entire data set into a regression function. In ICL, the regression operator $\Phi_n$ receives the data $D_n$ and a query $x_{q}$ as a {\em prompt}, and the goal is for $\Phi_n(D_n)(x_{q})$ to predict the value $f(x_{q})$ accurately.

In our study, following previous ICL investigations in regression, 
training employs noiseless data (but testing is performed on noisy data). 
Hence, the prompt at training time is:
\begin{equation}
   \overbrace{\underbrace{\underbrace{\v{x}_1, f(\v{x}_1)}_{\text{Example}_1}, \underbrace{\v{x}_2, f(\v{x}_2)}_{\text{Example}_2}, \dots, \underbrace{\v{x}_n, f(\v{x}_n)}_{\text{Example}_n}}_{\text{In-Context Examples}}, \v{x}_{q}}^{\text{Prompt}}.
    \label{eq:prompt}
\end{equation}
Training is performed on multiple prompts corresponding to different functions $f$ picked randomly from $\sH$. At testing time, the objective is to use a given user prompt consisting of a noisy test dataset $D^*_n =  \{(\bx^*_1,y^*_1),\ldots,(\bx^*_n,y^*_n)\}$ and a test point $\v{x}^*_{q}$ to infer $f(\v{x}^*_{q})$, {\em without any further training}.
Note that the target function $f$ {\em is not fixed} as in traditional regression but varies both at training and testing time. In addition, the ICL regression operator must be able to handle in-context examples of varying length~$n$. In sequential architectures, this is accomplished by processing the prompt autoregressively; however, this is not required and the prompt could be in principle processed in parallel in a more efficient manner, which is what we propose in this paper. Notice also that the problem is permutation-invariant to the arrangement of in-context examples. Hence, ICL in regression tasks is an example of {\em set learning}. 

\subsection{DeepONets}

A DeepONet consists of a branch network and a trunk network (see the right side of Fig.~\ref{fig:model}). The branch network takes the input $X$ and encodes this information into a feature vector $b_1,\ldots,b_p$. The trunk network, on the other hand, takes a query point $\v{x}$ at which the output is to be evaluated, and computes a feature vector $t_1,\ldots,t_p$. The final output of the DeepONet is obtained by taking the dot product of these feature vectors and adding a bias term:
\begin{equation}
    E(X;\v{w})(\v{x})\,=\, \sum_{i=1}^p b_i(X;\v{w}_{br}) \cdot t_i(\v{x};\v{w}_{tr}) + b_0\,.
    \label{eq:deeponet}
\end{equation}
where $\v{w} = \{\v{w}_{br},\v{w}_{tr},b_0\}$ comprise the neural network weights. This can be thought as a basis expansion approximation, where the trunk network computes adaptive basis functions, and the branch network computes the expansion coefficients. The bias term $b_0$, while not strictly necessary, often enhances performance. 

\subsection{The DeepOSets Architecture}

Before presenting the DeepOSets architecture, we prove a universal representation theorem that motivates it. A regression operator $\Phi_n$
is said to be {\em permutation-invariant} if, for any
permutation $\pi: \{1,\ldots,n\} \rt \{1,\ldots,n\}$, we have
$\Phi_n((x_1,y_1),\ldots,(x_n,y_n)) \,=\, \Phi_n((x _{\pi(1)},y _{\pi(1)}),\ldots,(x _{\pi(n)},y _{\pi(n)}))$.
A regression operator is said to be {\em continuously sum-decomposable}
through $\sZ$ if there exists a continuous function $\phi: \sX
\!\times\! \sY \rt \sZ$ and a continuous operator $\Gamma:\sZ \rt \sH$
such that 
\beq
  {\textstyle \Phi_n(D_n) \,=\, \Gamma\left( \sum_{(x,y)\in D_n} \phi(x,y)\right)}\,.
\label{eq-sumdec}
\eeq

The following result is the counterpart of Thm.~7 in
\cite{zaheer2017deep} and Thm.~2.9 in \cite{wagstaff2019limitations}.

\begin{theorem}
  A continuous operator $\Phi_n: (\sX \times \sY)^n
\!\rightarrow\!\sH$ is permutation-invariant if and only if it is
continuously sum-decomposable through $R^{\binom{n+d+p}{n}}$.
\end{theorem}

The proof of the previous theorem is given in the Appendix. This result says that one can decompose a permutation-invariant regression operator into two components: the first is a sum-pooling of identical functions $\phi$, while the second is an operator acting on the output of the first. This motivates us to use a DeepSets neural
network with sum-pooling (or, equivalently, average pooling) to
approximate the first component, and a neural operator approximator
for the second. Note, however, that this is true provided that the dimensionality of
the latent space must be at least $\binom{n+d+p}{n}$, which is large for large $n$, $d$, and $p$.

\begin{figure*}
    \centering
    \includegraphics[width=0.95\textwidth]{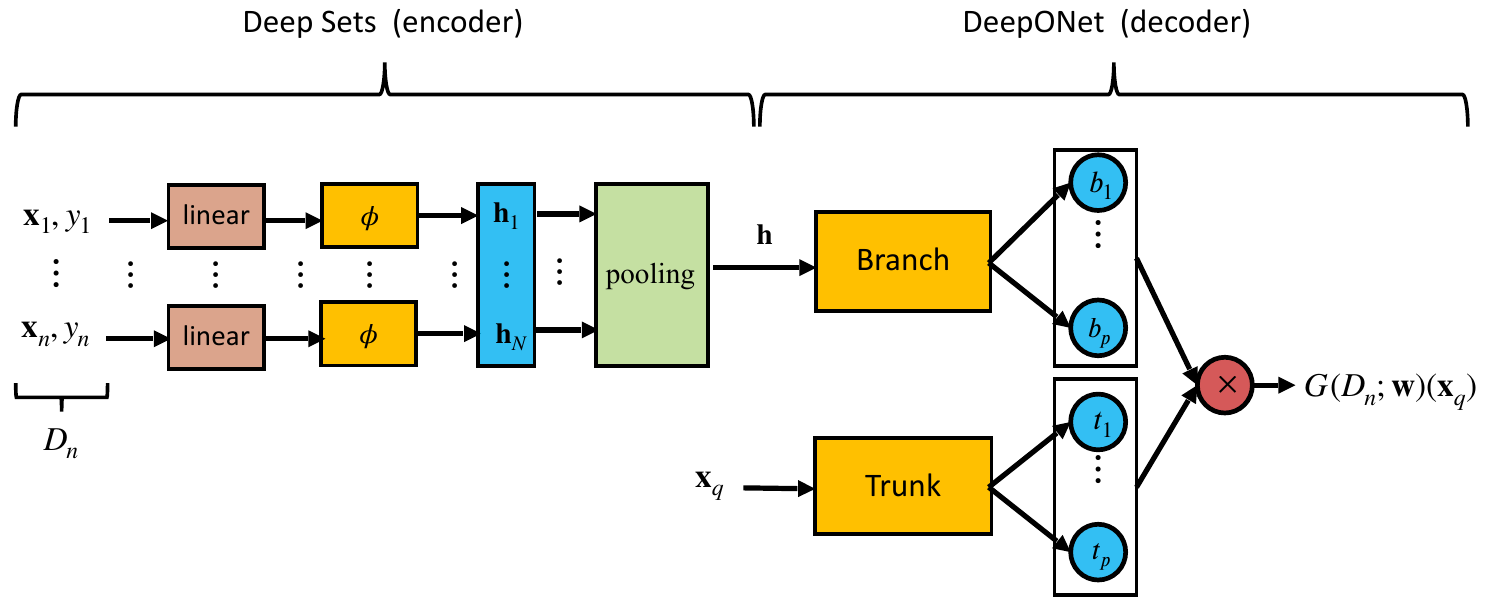}
    \caption{DeepOSets architecture for in-context learning of regression with built-in permutation invariance inductive bias.} 
    \label{fig:model}
\end{figure*}

For the neural approximator, we choose DeepONets, due to its excellent properties,
such as the fact that it has a universal representation theorem and
employs an implicit representation for the output function, requiring
no output discretization. However, DeepONets are not
directly usable for our purposes, due to the following two major
challenges: (1) the branch network accepts a fixed number of inputs,
and thus it cannot accommodate a varying number of in-context
examples; and (2) the branch network is sensitive to permutations of its input, which can lead to poor generalization since the data in regression is invariant to permutation. 

Using a DeepSets encoder before the branch network addresses both of
these challenges, allowing the DeepONet to process an indefinite
number of in-context examples in a permutation-invariant setting. This
leads to the DeepOSets architecture, displayed in Fig.~\ref{fig:model}.
Each in-context example $(\v{x}_i, y_i)  \in \mathbb{R}^{d+1}$ is embedded into a higher-dimensional space by means of a trainable linear layer, and then encoded into a latent space vector $\v{h}_i \in \mathbb{R}^{d_{embed}}$ by means of an MLP $\phi$. The encoded vectors $\v{h}_i$ are pooled into a single vector $\v{h}$ to obtain permutation invariance (details in Appendix). 
The simplest pooling method is to average across all features: $\v{h} = \frac{1}{n}\sum_{i=1}^{n} \v{h}_i$. The aggregated vector $\v{h}$ provides the input to the branch network of a DeepONet. Notice that the weights in the linear layer and MLP $\phi$ are shared by all inputs (indeed, the linear layer can be seen as part of $\phi$). The permutation invariance property of DeepOSets with respect to the in-context examples provides an inductive bias that can improve the model generalization ability.

The next result, the proof of which can be found in the Appendix, shows that DeepOSets are dense in the space of continuous permutation-invariant operators.

\begin{theorem} Let $\Phi_n:(\sX \times \sY)^n \rt \sH$ be a
continuous permutation-invariant operator, and suppose that the
hypothesis space $\sH$ is $C(K)$, the space of continuous real-valued
functions on a compact subset $K \subseteq R^p$.  For any $\eps>0$,
there are weights $\v{w}$ and a DeepOSets neural network
$G(\cdot;\v{w})$, such that
\beq
  \left|G(D_n;\v{w})(\v{x}_{q}) - \Phi_n(D_n)(\v{x}_{q})\right| \, < \, \eps\,,
\eeq
for all $D_n \in (\sX \times \sY)^n$ and $\v{x}_q \in K$.
\end{theorem}

Two variants of this basic architecture are considered. {\em
  DeepOSets-T} replaces the DeepSets layer by a {\em Set Transformer}
\citep{lee2019set}. We have found empirically that this improves
accuracy in high-dimensional cases. In order to mitigate the quadratic
complexity of the set transformer, {\em DeepOSets-TI} utilizes inducing
points, as described in~\cite{lee2019set}.

As we see in the next section, DeepOSets can perform in-context
learning with a smaller number of weights than transformer-based
approaches. Moreover, the baseline DeepOSets achieves linear training
complexity $\mathcal{O}(n)$ in the number of in-context examples. At
inference time, after the first query, all subsequent queries, with
fixed in-context examples, have constant complexity $\mathcal{O}(1)$,
as the prompt does not need to be processed again. In contrast,
transformer approaches, which include DeepOSets-T and the transformer
approach in \cite{garg2023}, have complexity $O(n^2)$ during
training. However, at inference time, DeepOSets-T has complexity $O(1)$ after the first query, while a standard transformer approach still
has complexity $O(n^2)$, as it requires recomputation of attention
matrices over the entire prompt sequence for each new
query~\citep{keles2023computational}.  As each query $x_q$ is appended
to the prompt, the attention computation must be redone even if the
rest of the prompt remains unchanged, incurring quadratic cost per
query. Key-value (KV) caching~\citep{hooper2024kvquant, ge2023model} can reduce this to
$\mathcal{O}(n)$ by reusing computations for the fixed portion of the
prompt.

\section{Results}

In this section, we use a set of stylized regression tasks, including
linear, polynomial, and neural network regression in low and high dimensions, to
demonstrate the in-context learning ability of DeepOSets and compare
it to a popular transformer-based method~\cite{garg2023}. 
Performance is assessed in terms
of accuracy, architecture size (number of weights), and
computational complexity. DeepOSets were implemented in JAX~\citep{jax2018github} and
Equinox~\citep{kidger2021equinox}, and other variations were
implemented in PyTorch \citep{paszke2017automatic}. For the
transformer component, we used the publicly available source code
from~\citep{garg2023}. Visualization of results was performed with
Seaborn~\citep{Waskom2021} and Matplotlib~\citep{Hunter:2007}. All
experiments were conducted on Nvidia A100 GPUs. 
 
\subsection{Linear Regression}
\label{sec:train}

In linear regression, the hypothesis space consists of linear functions
\begin{equation}
    \mathcal{H} = \{f | f(\v{x}) = \v{w}^T\v{x},\, \v{w} \in \mathbb{R}^d\},
\end{equation}  
where $\v{w}$ is randomly picked to generate the training data. Following~\cite{garg2023}, we let $\v{w} \sim \mathcal{N}(0, \v{I}_d)$. Given a sample function $f$ thus generated, we again follow~\cite{garg2023} and generate each context example $\v{x}_i$ from $\mathcal{N}(0, I_d)$ and obtain the corresponding target $f(\v{x}_i)$. 

\begin{figure}
    \centering
    \includegraphics[width=0.7\textwidth]{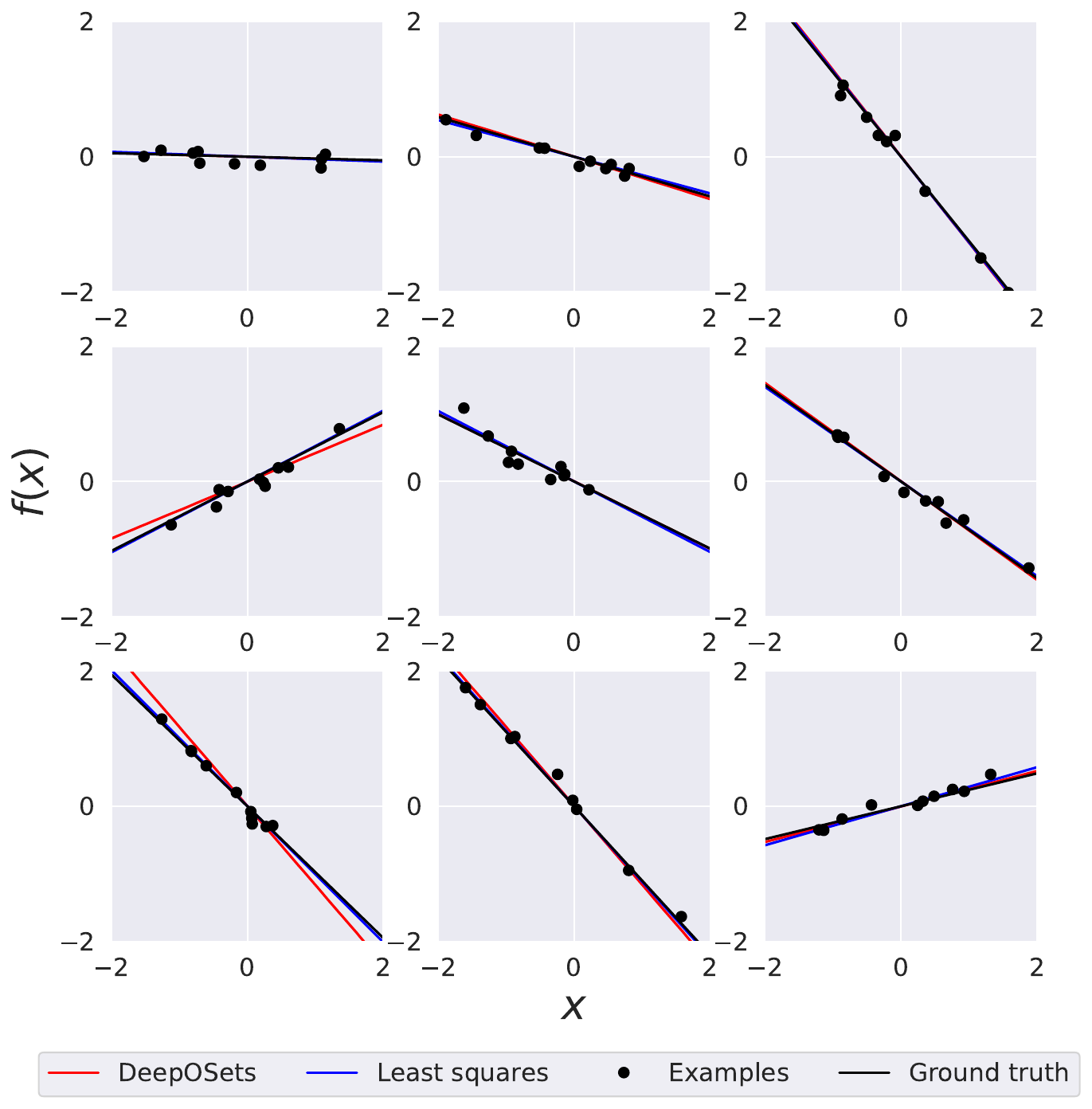}
    \caption{Learning linear regression with DeepOSets with $n=13$
      in-context examples in the training set. The black dots
      represent 10 in-context test examples corrupted by Gaussian noise $\epsilon \sim \mathcal{N}(0, \sigma^2 = 0.1)$.}
    \label{fig:training}
\end{figure}

Fig.~\ref{fig:training} displays several examples of training data sets
and linear regression obtained by DeepOSets in
the case $d=1$. The model was trained on noiseless prompts of size
$n=13$. Training converged in 9 minutes with 16K iterations and reached training mean square error 7.9e-4. At test time, the model was challenged with prompt examples corrupted by noise: $\tilde{f}(\v{x}) = \v{w}^T\v{x} + \epsilon$, where $\epsilon \sim \mathcal{N}(0, \sigma^2)$, where $\sigma^2=0.1$. In addition, the sample size at inference time is $n=10$, thus different than the sample size used for training. We can see in Fig.~\ref{fig:training} that the trained DeepOSets model accurately recovers the ground-truth function. The ordinary least-squares regression line is also displayed, for comparison. 

We also considered more realistic high-dimensional ($d=20$) linear
regression. While the original DeepOSets performs very well with
$d=1$, we found that this is not the case with $d=20$.  This issue
likely stems from the limited capacity of the DeepSets layer to
capture higher-order interactions. To address this, we replace
DeepSets with the Set Transformer architecture~\citep{lee2019set},
which leverages both attention mechanisms and permutation-invariant
inductive biases. Additionally, we incorporate residual connections
into both the Set Transformer and DeepONet components to stabilize
training in deeper models by mitigating vanishing or exploding
gradients. DeepOSets-T produced very accurate results in the $d=20$
case, at the expense of increasing the training complexity from $O(n)$
to $O(n^2)$. To mitigate this issue, we employ the inducing points
method introduced by~\cite{lee2019set} to reduce both time and memory
complexity. In this approach, a set of trainable inducing points
serves as queries within the self-attention block, enabling the model
to encode the input set more efficiently. The resulting DeepOSets-TI
architecture reduces the time and space complexity of the attention computation from
$O(n^2)$ to $O(nm)$, where $n$ is the number of input elements and $m$
is the number of inducing points.

We use the transformer method in \cite{garg2023} as a baseline for comparison in our quantitative experiments. In all cases, we apply 500K training steps, learning rate 1e-4 with a cosine annealing scheduler. For the transformer, we remove the positional encoding to improve generalization. During the training steps of DeepOSets-T, we uniformly sampled the data size between 21 to 41, but tested on much longer prompts with up to 500 examples.
Training of DeepOSets-T took 3 hours, which is significantly faster
than the transformer baseline, which took 11 hours to train. We
evaluate the effect of different values of $m$ in the inducing point
method, namely $m=21$, and $m=41$. Please see the Appendix for a full
description of the hyperparameters used in the simulations.

The results are displayed in Tab.~\ref{tbl:inference}. DeepOSets achieves superior performance with significantly fewer parameters compared to the transformer baseline, particularly for $d=20$. The transformer exhibits degraded performance when the number of in-context examples exceeds the size used during training, implying a lack of robustness to distribution shifts. In contrast, DeepOSets-T remains stable across varying in-context set sizes.
While the transformer required 11 hours to complete training, DeepOSets-T achieved convergence in only 3 hours. Notably, the inclusion of inducing points in DeepOSets-TI does not lead to a significant increase in training time (Tab.~\ref{tbl:deeposet-ts}).
DeepOSets-T shares the same complexity as transformers for processing the first query but avoids computing full attention matrices. Instead, it encodes the in-context set into a latent representation, enabling constant-time inference per query. Although DeepOSets-TI shows a minor drop in accuracy on noiseless linear regression, it maintains robustness and accuracy on larger datasets where the transformer struggled.

\begin{table*}[tb]
    \small
    \centering 
    \caption{Linear regression benchmark with $d=20$, $n=41, 500$ with different levels of noise where $n$ is the testing sample size for DeepOSets. TF is the transformer from \cite{garg2023}. $m$ represents the number of inducing points used in Set Transformer.}  
    \label{tbl:inference}
    \begin{tabular}{lccccc}
    \toprule
         & TF &DeepOSets-T &DeepOSets-TI(m=21)&DeepOSets-TI(m=41) & OLS\\
    \hline
    \# of param. &9.5M&1.8M&2.2M&2.2M&0\\
    Training time &11hr &3hr&3hr&3hr& 0\\
    \hline
    LR ($\sigma^2=0, n=41$) &4.79e-3  & 1.73e-2&1.24e-1&1.139e-1 &1.98e-12\\
    LR ($\sigma^2=0, n=500$) & 1.508  &7.23e-3&2.55e-2&1.725e-2 & 8.377-13\\
    \hline
    LR ($\sigma^2=1, n=41$) & 2.47  &5.4&2.839&3.564&1.88\\
    LR ($\sigma^2=1, n=500$) & 2.92 &3.65&1.179&3.085&9.17e-1\\
    \hline
    LR ($\sigma^2=2, n=41$) & 7.65 &9.012&8.628&1.641e1&6.1\\
    LR ($\sigma^2=2, n=500$) &7.35 &5.65&3.455&3.821& 3.657\\
    \hline
    \end{tabular}
    \label{tbl:deeposet-ts}
\end{table*}

\begin{figure*}[tb]
    \centering 
    \begin{subfigure}[t]{0.5\textwidth}
        \caption{linear regression ($\sigma^2=0.04$)}
        \includegraphics[width=\textwidth]{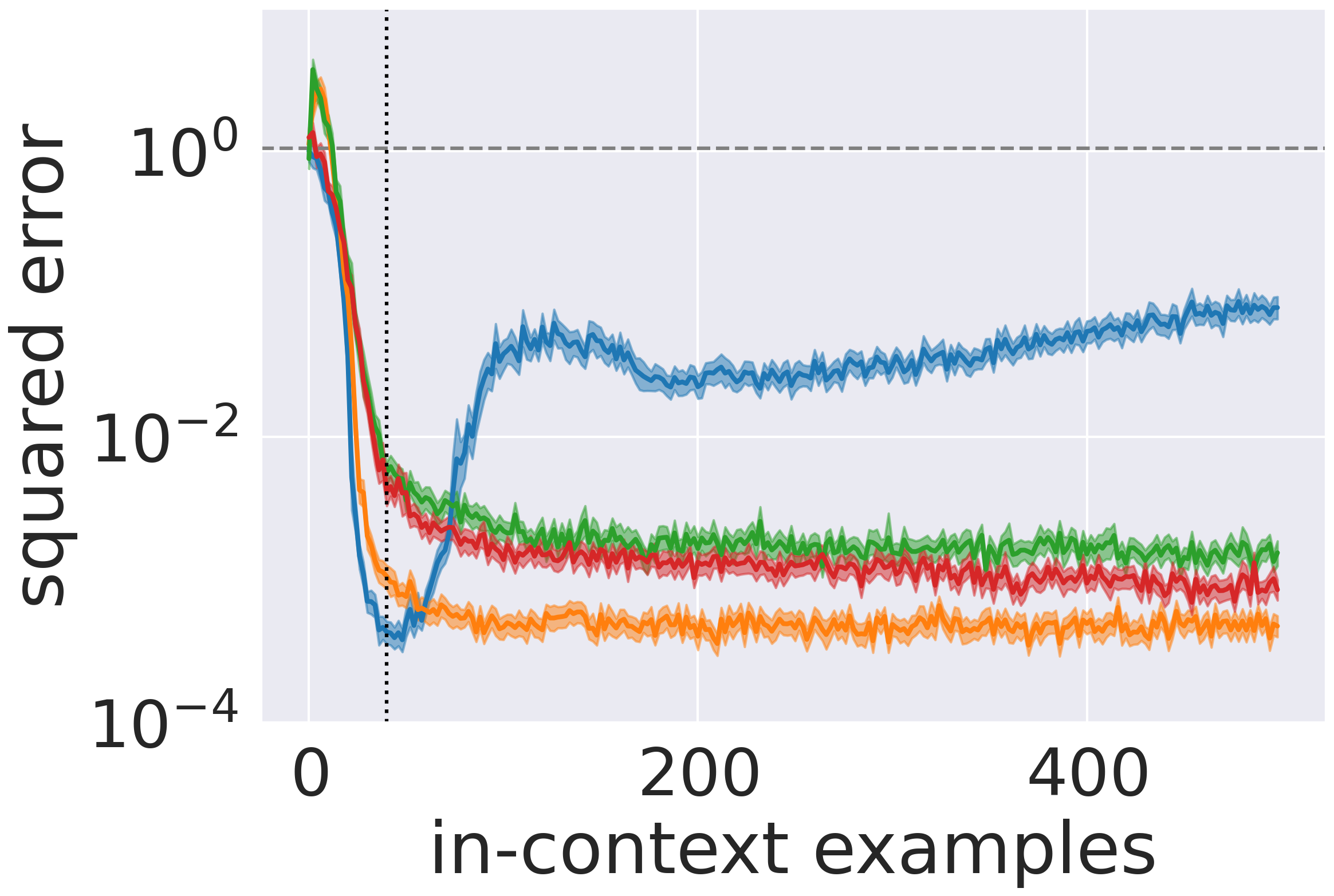}
    \end{subfigure}%
    \begin{subfigure}[t]{0.5\textwidth}
        \caption{linear regression ($\sigma^2=0.2$)}
        \includegraphics[width=\textwidth]{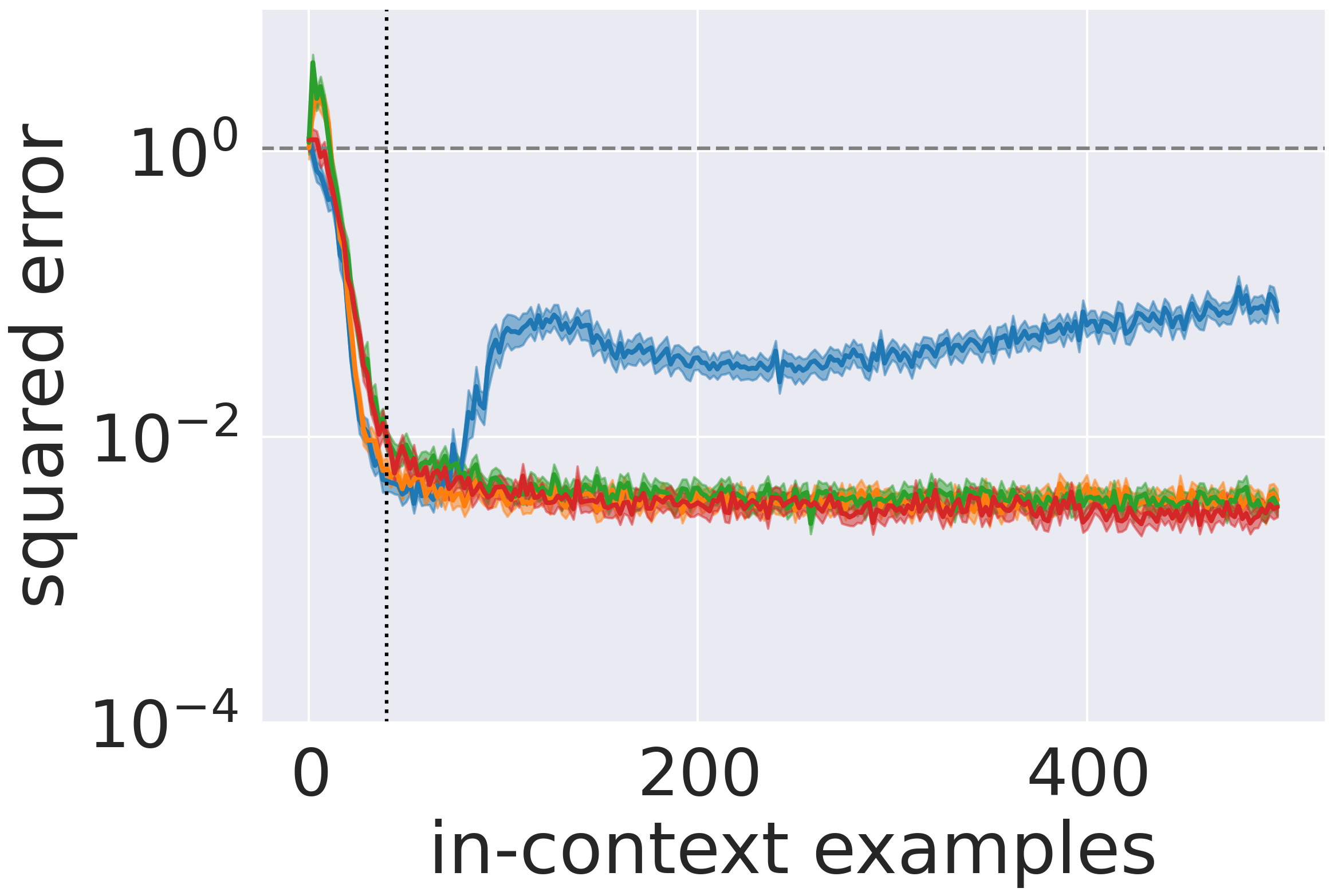}
    \end{subfigure}

    \begin{subfigure}[t]{\textwidth}
        \centering
        \includegraphics[width=0.95\textwidth]{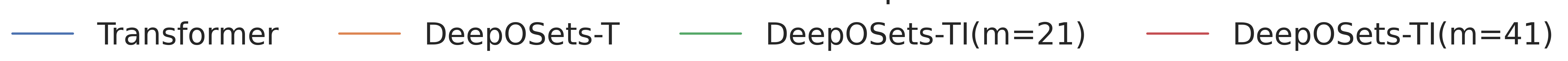}
    \end{subfigure}
     \caption{Performance comparison of DeepOSets-T (with a Set Transformer), its variant with inducing points (DeepOSets-TI), and a transformer baseline on 20-dimensional linear regression. The vertical line denotes the training set size of in-context examples (41), and $m$ indicates the number of inducing points used in DeepOSets-TI.}
     \label{fig:train-lr-ind}
\end{figure*}

Fig.~\ref{fig:train-lr-ind} examines in more detail the performance of
DeepOSets-T against its variant with inducing points DeepOSets-TI and
a transformer baseline on 20-dimensional linear regression. We can see
that DeepOSets-TI produces essentially no loss of accuracy with
respect in the high-noise case ($\sigma^2 = 0.2)$, and that both
DeepOSets-T and DeepOSets-TI are more accurate than the baseline
transformer as the number of test examples increases. We hypothesize
that this is due to the difficulty the baseline transformer
architecture introduces in sequential processing of long prompts.
Overall, the use of inducing points offers a practical trade-off
between efficiency and performance, particularly in scenarios where
memory or runtime constraints are critical.

\subsection{Neural Network Regression}

\begin{figure*}[tb]
    \centering
    \begin{subfigure}[b]{0.5\textwidth}
        \caption{neural network regression (noiseless test data)}
        \includegraphics[width=\textwidth]{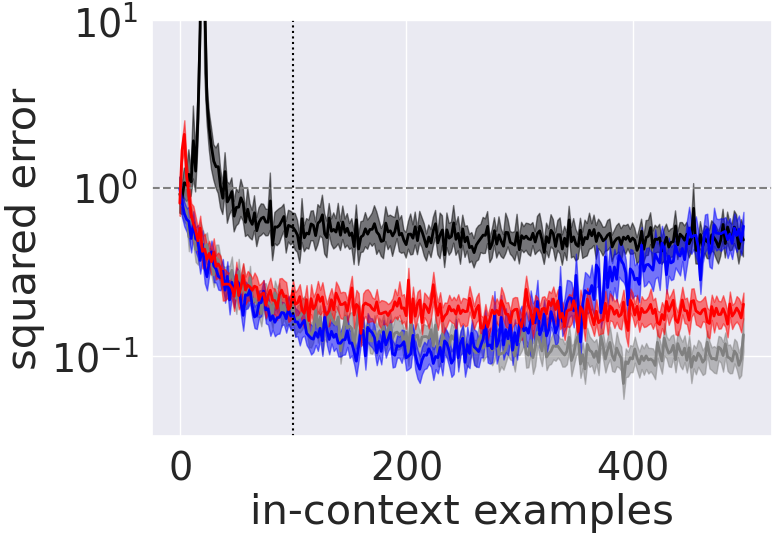}
        \label{fig:train-relu-test-relu}
    \end{subfigure}%
    \begin{subfigure}[b]{0.5\textwidth} 
        \caption{linear regression (noisy test data, $\sigma^2 = 1$)}
        \includegraphics[width=\textwidth]{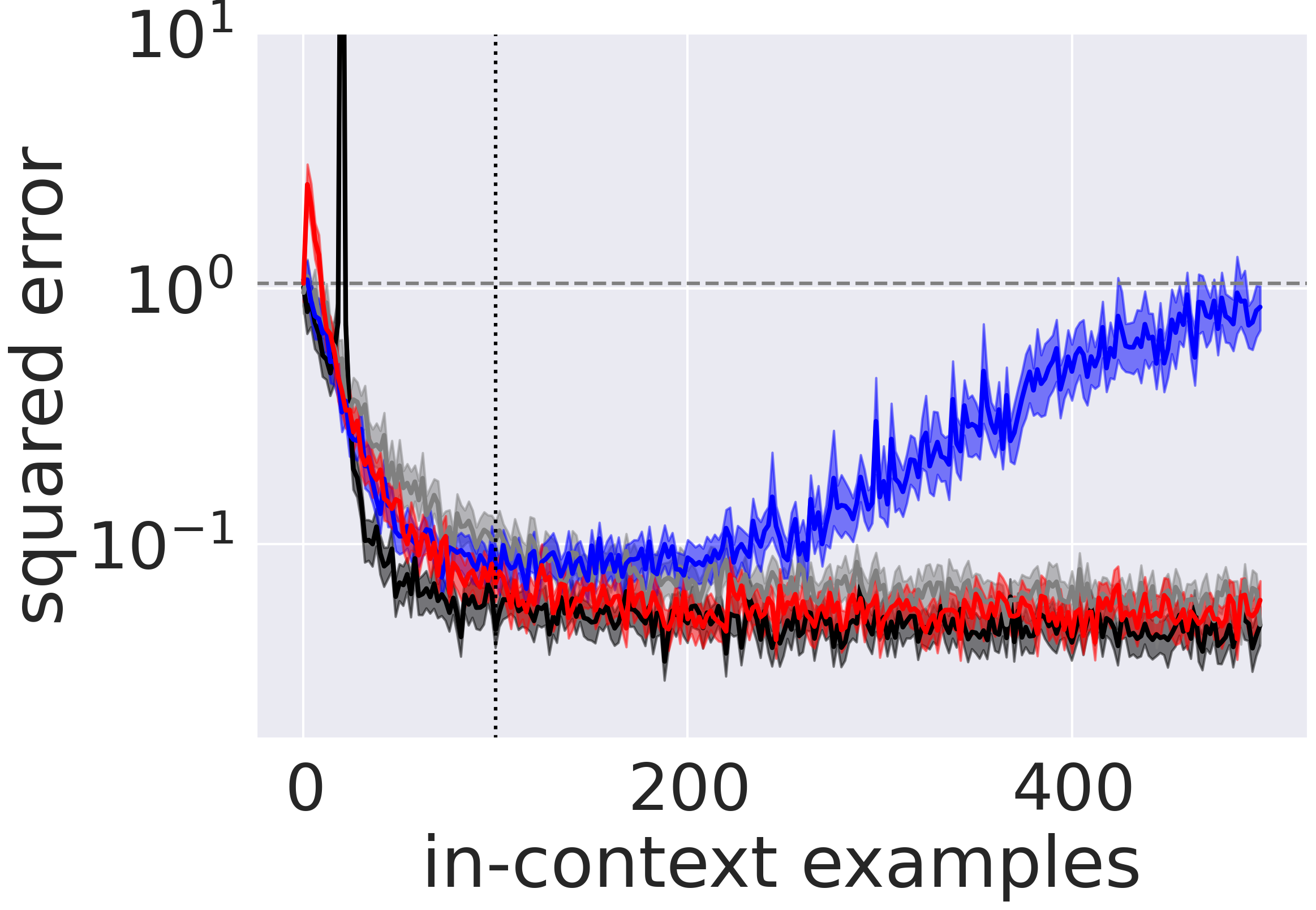}
        \label{fig:train-relu-test-noisylr}
    \end{subfigure}%

    \begin{subfigure}[b]{0.9\textwidth}
        \includegraphics[width=\textwidth]{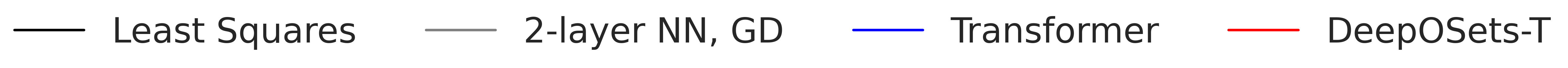}
    \end{subfigure}
   
    \caption{(a) Performance of DeepOSets-T on 20-dimensional shallow neural network regression. The vertical line indicates the size of the training set (101). (b) The same model also performs well on linear regression tasks, achieving results comparable to standard neural network training. These experiments further highlight the consistency of DeepOSets-T when handling long in-context sequences.}
    \label{fig:train-relu}
\end{figure*}

We evaluated the ability of the DeepOSets-Transformer architecture to perform ICL in non-linear regression by using as hypothesis space the set of two-layered neural network with ReLU activation function. This particular function class was chosen because it represents a common and challenging form of non-linearity. The
hypothesis space here is
\begin{equation}
    \mathcal{H} = \{f | f(\v{x}) = \sum_{i=1}^{r} c_i
    \sigma(\v{w}_i^\top \v{x}),\, \v{w}^\top_i \in \mathbb{R}^d, c_i
    \in \mathbb{R}\}\,.
  \end{equation}
In our experiments, we consider $r=100$ hidden units, ReLU
activation $\sigma(x)=\max(0,x)$, $d=20$, and randomly generated network weights $c_i$ and $\v{W}_i$ from $\mathcal{N}(0,2/r)$ and $\mathcal{N}(0, I_d)$. Fig.~\ref{fig:train-relu-test-relu} shows that DeepOSets-T is
capable of in-context learning for this function class. Its
prediction error decreases at a rate comparable to a baseline that trains a neural network via gradient descent.
To assess generalization, Fig.~\ref{fig:train-relu-test-noisylr} evaluates the trained model on linear regression tasks with additive noise. Despite being trained on neural network data, DeepOSets-T demonstrates robust predictive performance in all evaluated settings.  This setup presents a significant out-of-distribution challenge, as the model was exclusively trained on data generated from two-layer ReLU neural networks, a non-linear function class. The results consistently demonstrate that DeepOSets-T exhibits robust and effective predictive performance across all evaluated settings, despite this task-domain shift. This strong performance is particularly notable because the model was not explicitly trained to handle linear functions or noise.

\subsection{Polynomial Regression with Automatic Model Selection}

The goal of this experiment is to investigate whether a trained DeepOSets can
perform optimal model selection, i.e., find the optimal order of ploynomial regression for a given data set provided in the prompt.
The hypothesis space consists of polynomials of order ranging from 1 to 10:
\begin{equation}
    {\textstyle \mathcal{H} = \{f | f(x) = \sum_{i=1}^{r} c_i x^{i}, c_i \in \mathbb{R}, r=1,\ldots,10\}}\,,
\end{equation}  
where the coefficients $c_i$ and $r$ are randomly picked to generate
the training data. We assume that $c_i = d_i e^{-i}$, where $d_i \sim
\mathcal{N}(0,1)$ --- the exponential dampening for higher-order terms
is introduced to reduce excessive oscillation. The polynomial order
$r$ is uniformly sampled from $\{1, \dots, 10\}$. Performance of
DeepOSets is compared to the traditional least-squares polynomial regression with
order selected by leave-one-out cross-validation, which we call
``Polyfit''.

\begin{figure*}[tb]
    \centering
    \begin{subfigure}[t]{0.4\textwidth}
        \caption{low order ($r=3$, $n=30$)}
        \includegraphics[width=\textwidth]{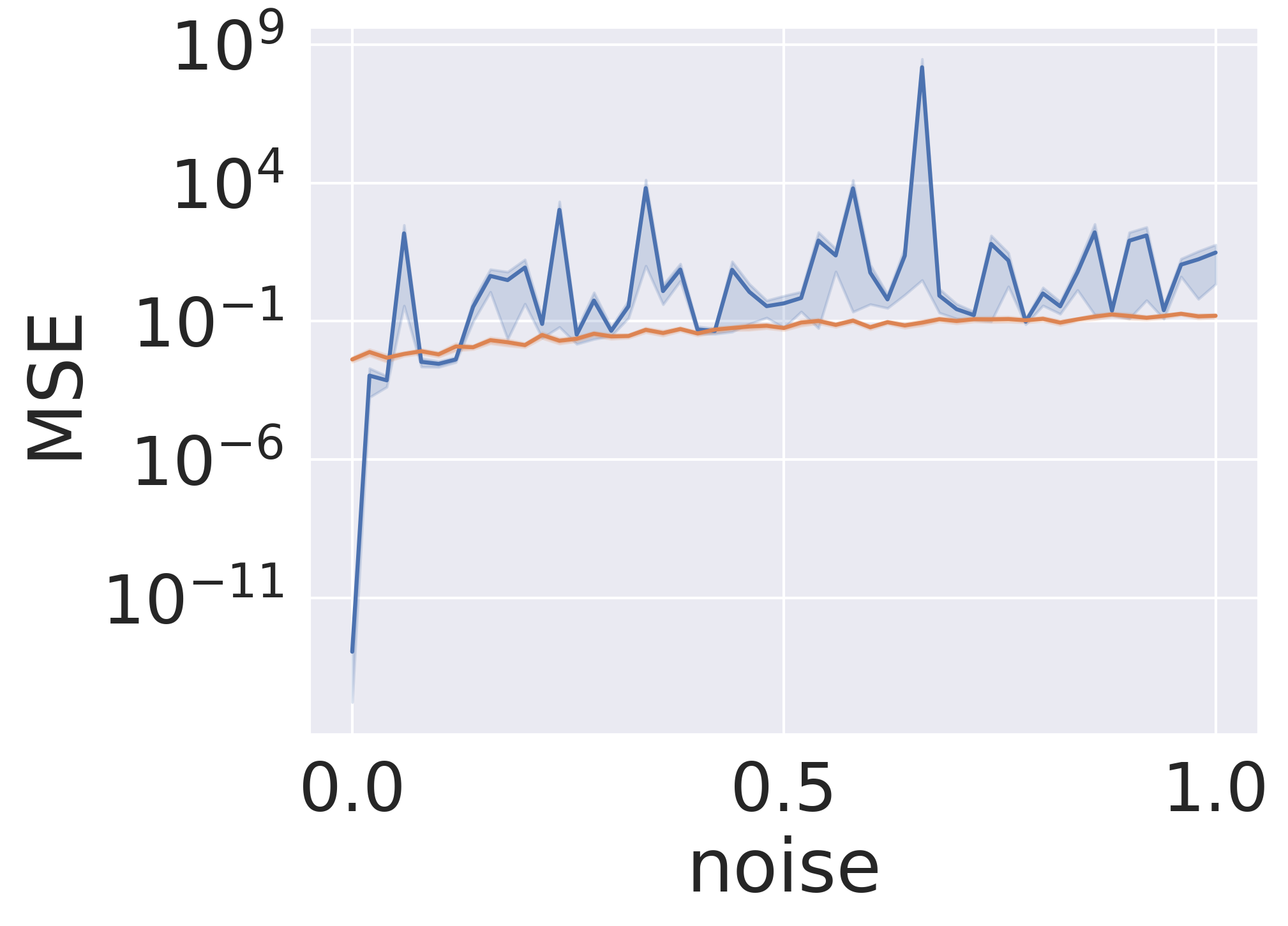}
        \label{fig:poly-low-k-noise-sweep}
    \end{subfigure}%
    \begin{subfigure}[t]{0.4\textwidth}
        \caption{high order ($r=7$, $n=30$)}
        \includegraphics[width=\textwidth]{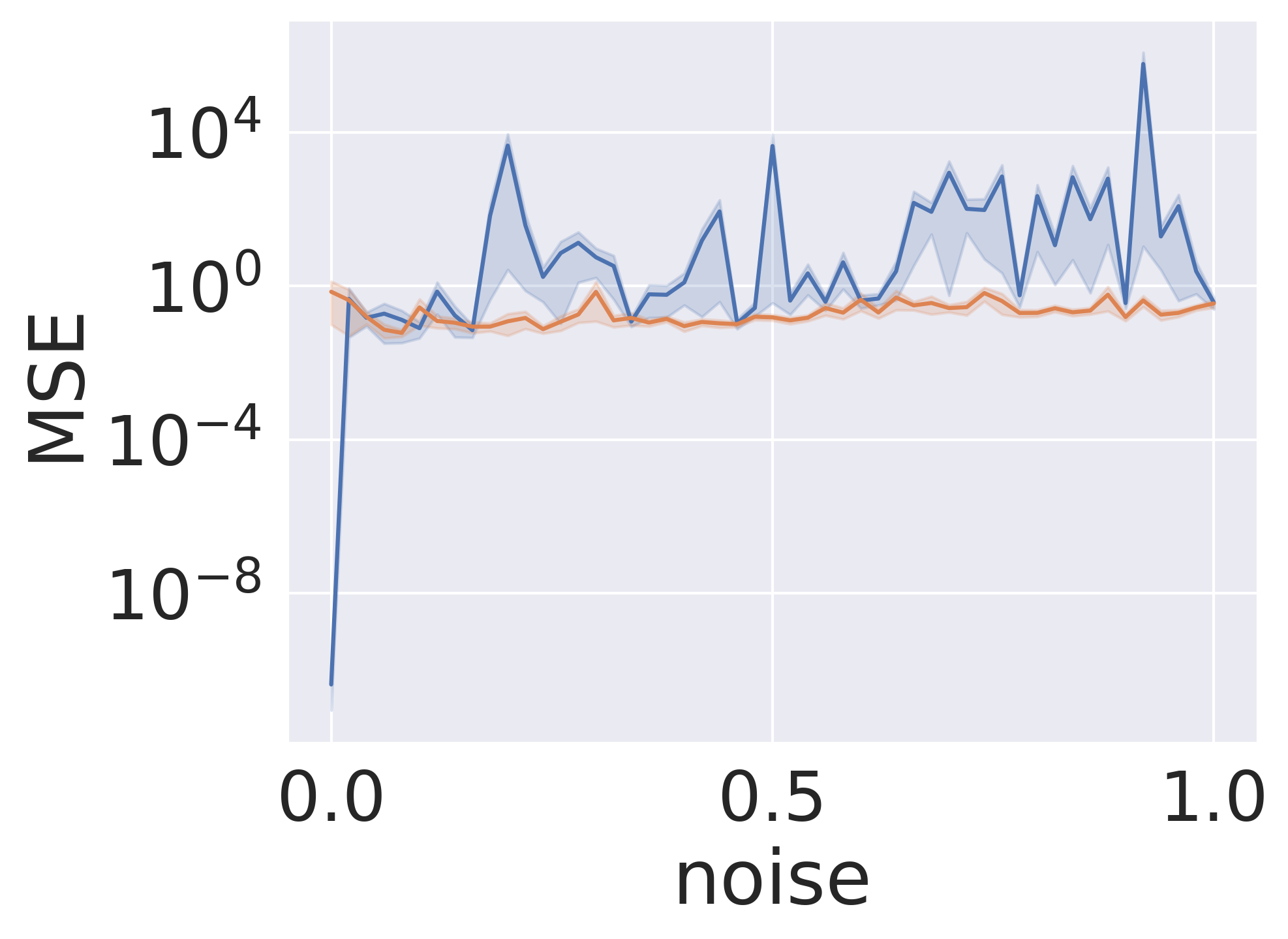}
        \label{fig:poly-high-k-noise-sweep}
    \end{subfigure}\\
    \begin{subfigure}[t]{0.4\textwidth}
        \caption{low order ($r=3$, $\sigma^2 = 0.1$)}
        \includegraphics[width=\textwidth]{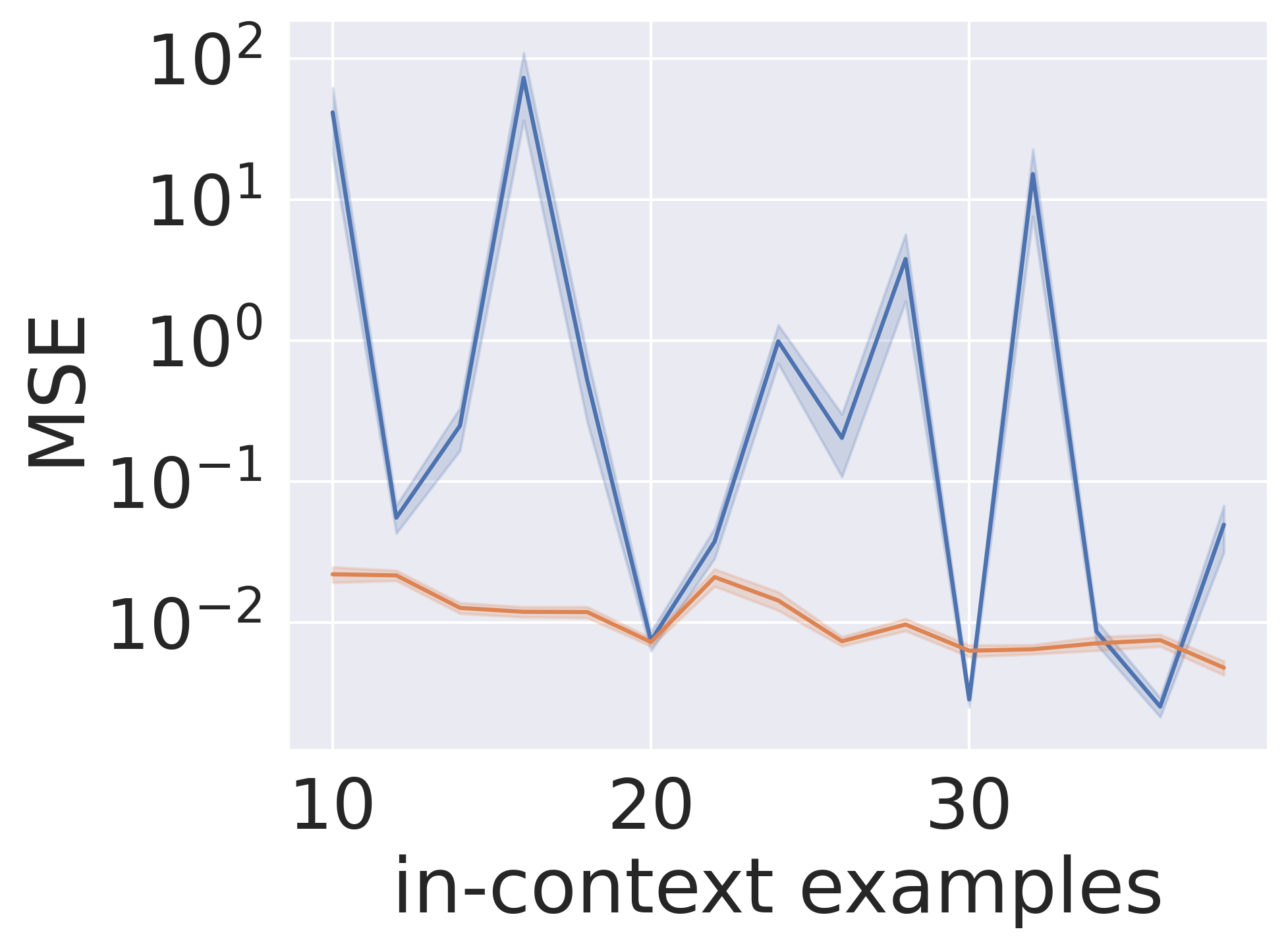}
        \label{fig:poly-low-k-exp-sweep}
    \end{subfigure}%
    \begin{subfigure}[t]{0.4\textwidth}
        \caption{high order ($r=7$, $\sigma^2 = 0.1$)}
        \includegraphics[width=\textwidth]{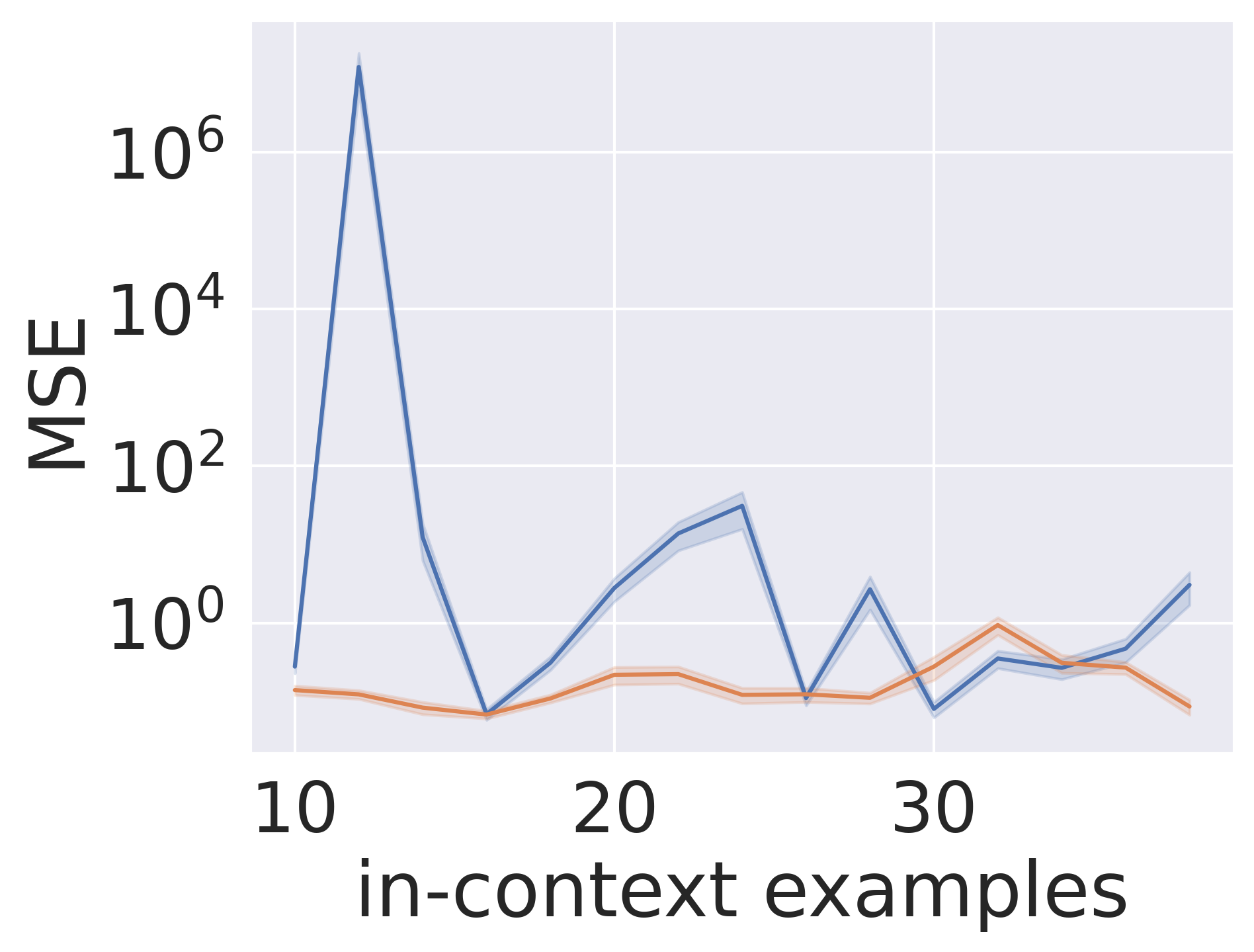}
        \label{fig:poly-high-k-exp-sweep}
    \end{subfigure}
 
    \caption{Polynomial regression results: The blue line (\textbf{\textcolor{blue}{---}}) represents the leave-one-out polyfit baseline, while the orange line (\textbf{\textcolor{orange}{---}}) depicts the DeepOSets model. Results are shown across 30 randomly generated functions.}
\end{figure*}

The training data is generated with polynomials of order $r=1$ to $r=10$, while the test data corresponds to either $r=3$ or $r=7$. We computed the mean squared error (MSE) as a function of noise level, with 15 in-context examples (Fig.~\ref{fig:poly-low-k-noise-sweep} and Fig.~\ref{fig:poly-high-k-noise-sweep}). The results indicate that polyfit, which performs leave-one-out for model selection, exhibits higher variance compared to DeepOSets, especially in high-noise regimes. We can see that the performance of DeepOSets remains stable across different noise levels.  
DeepOSets demonstrates robustness across different numbers of in-context examples, maintaining consistently lower error (Fig.~\ref{fig:poly-low-k-exp-sweep} and Fig.~\ref{fig:poly-high-k-exp-sweep}). In contrast, polyfit is computationally intensive and shows greater sensitivity to the number of examples and larger errors (Tab.~\ref{tbl:inference-poly}).

\begin{table}[b!]
    \centering 
    \caption{Polynomial regression with r=7, n=30, and $\sigma^2=0.2$.}
    \begin{tabular}{lcc}
    \toprule
         &  DeepOSets & Polyfit\\
      \midrule
      Parameters & 72K & \textbf{0} \\
      Training time & 2min & \textbf{0}\\ 
      Inference complexity & $\boldsymbol{O(n)}$ & $O(n^3)$\\
      Inference time & \textbf{12 ms} & 38 ms\\
      Test MSE  & \textbf{0.08} &  1.68\\
      \bottomrule
    \end{tabular}
    \label{tbl:inference-poly}
\end{table}

\section{Conclusion}

We investigated in this paper the emergence of ICL in a non-autoregressive, permutation-invariant neural architecture, called DeepOSets. 
This architecture combines set learning and operator learning in an innovative way, and may be of independent interest in both of these tasks. 
We proved that DeepOSets is a universal approximator for permutation-invariant
regression operators. 
Experimental results with linear and shallow neural network regression
have unveiled the potential of DeepOSets as a more efficient and
noise-robust alternative to autoregressive models. 
Several key findings regarding the performance and
behavior of DeepOSets were observed. The DeepSets module effectively
generalizes in-context examples, enabling DeepOSets to maintain
consistent performance with noisy prompts of varying size not
encountered during training. The DeepOSets model is trained
with noiseless data, and the ability to predict noisy prompts is
meta-learned. The experiments with polynomial regression showed that DeepOSets can
learn multiple operators and perform automatic model selection. This
shows that DeepOSets has potential as an auto-ML approach, where a single network can be trained with diverse data
and be able to automatically select the hyperparameter from a new prompt, with no further training. Future work will continue to investigate the properties of the DeepOSets architecture in in-context learning, set learning, and operator learning tasks.

\section{Acknowledgements}

This material is based upon work supported by the National Science Foundation under Grant Number 2225507. The authors are grateful to Jonathan Siegel for his comments on the proof of Theorem 1.

\section{Appendix}

{\bf Proof of Theorem 1.}  The reverse direction follows immediately, since the composition of two continuous mappings is continuous and the right-hand side of
\eqref{eq-sumdec} is clearly permutation-invariant.
The proof of the forward direction similar to that in the proof of Theorem~7
in \cite{zaheer2017deep} for permutation-invariant functions, but here
it is extended to functions operating on vectors. For notational simplicity, we
will denote the data $D_n$ as a matrix $U = (\v{u}_1,\ldots\v{u}_{n})^T
\in R^{n\times m}$, where $\v{u}_i = (\v{x}_i,\v{y}_i) =
(u_{i1},\ldots,u_{im})$ is the $i$th data point, with $m = d+p$.
Because the operator $\Psi_n$ is permutation-invariant, we may order
the data points (the rows of $U$) using the lexicographical order
$u_i <_{\rm lex} u_j$ if  $u_{ik}<u_{jk}$ for the smallest
$k=1,\ldots,m$ where $u_{ik} \neq u_{jk}$ and, by scaling the data if
needed, assume that the data is an element of the following subset of $R^{n\times m}$:
\beq
 \sU \,=\, \{(\v{u}_1,\ldots\v{u}_{n})^T \in [0,1]^{n\times m} \mid u_1 \leq_{\rm lex} u_2 \leq_{\rm lex} \cdots \leq_{\rm lex} u_n\}\,.
\eeq

In \cite{zaheer2017deep}, the Newton identities relating power
sums to the elementary symmetric polynomials were used to construct an
injective map from $\sU$ (in the case $d=1$) to a compact latent space
$\sZ \subset R^{n+1}$, which was the key part of their proof. This
argument can be extended to the case $d>1$ by using the {\em elementary
multisymmetric polynomials}, but the dimensionality of the latent space
needs to be much larger than $n+1$. Given a multi-index $\vv{\alpha} =
(\alpha_1,\ldots,\alpha_m) \in N^m$ and $u_i \in [0,1]^m$, let $u_i^{\vv{\alpha}} =
u_{i1}^{\alpha_1}u_{i2}^{\alpha_2}\cdots u_{im}^{\alpha_m}$. Now
consider the set $\{u_i^{\vv{\alpha}} \mid |\vv{\alpha}| = \alpha_1+\cdots\alpha_m
\leq n\}$ of all multivariate polynomials $u_i^{\vv{\alpha}}$ of order up to
$n$, which has cardinality $t = \binom{n+m}{n}$. Consider an
enumeration of these polynomials $\{u_i^{\vv{\alpha}_1},\ldots,
u_i^{\vv{\alpha}_t}\}$. With $m=1$ and $u_{i1}=x$, this is just the set 
$\{1,x,x^2,\ldots,x^{n+1}\}$ in
\cite{zaheer2017deep}. With $n=3$, $m=2$, $u_{i1}=x$, $u_{i2}=y$ we
would get the set $\{1,x,y,xy,x^2,y^2,xy^2,x^2y,x^3,y^3\}$.
Now let the continuous function $\psi: [0,1]^m \rt R^t$ be specified by
each component $\psi_k(u_i) = u_i^{\vv{\alpha}_k}$, $k = 1,\ldots,t$ and consider the
map $E: \sU \rt R^t$, where each component is a {\em power-sum multi-symmetric polynomial}:
\beq
E_k(U) \,=\, \sum_{i=1}^n \v{u}_i^{\vv{\alpha}_k}\,, \quad k=1,\ldots,t.
\eeq
Note that $E(U) \,=\, \sum_{i=1}^n \psi(u_i)$. Now, Proposition~1 of
\cite{maron2019provably} states that $E(U) = E(V)$ if and only if $V$
is a permutation of the rows of $U$. This implies that if $U,V \in
\sU$, then in fact $E(U) = E(V)$ if and only if $U = V$, in other
words, $E$ is injective on $\sU$. Let $\sZ = E(\sU) \subset R^{t}$ be the range
of $E$, which is compact since $\sU$ is compact and $E$ is
continuous. Then $E$ is a bijection between $\sU$ and $\sZ$ and, furthermore,
$E$ is continuous with a continue inverse $E^{-1}$, since $E$ is a
polynomial map between compacts $\sU$ and $\sZ = E(\sU)$.
Now, define the operator $\Gamma:R^{t}\rt \sH$ via $\Gamma = \Psi_n \circ E^{-1}$. As a composition of
continuous mappings, the operator $\Gamma$ is continuous, with
$\Psi_n(U) = \Gamma(E(U)) = \Gamma(\sum_{i=1}^n \psi(u_i))$, and the theorem is proved. 

\bigskip

\noindent{\bf Hyperparameter Settings}
\paragraph{DeepOSets} With $d=1$, we linearly embed $\v{x}$ and $f(\v{x})$ into 5-dimensional space. Then, the embedded examples are processed by a 6-layer MLP with 50 hidden units into a vector of size 400. Both branch and trunk nets contain 5 layers with 40 hidden units. The last layer of DeepOnet contains 100 units. This resulted in a total number of trainable parameters equal to 72K. Following~\cite{zaheer2017deep}, we employed the SELU activation function~\cite{klambauer2017self} in the DeepSets module. For the trunk network, the tanh nonlinearity is used. Training employs the Adam optimizer~\cite{kingma2014adam} with a learning rate of 1e-3 and exponential decay by 0.9 every 2000 steps. 

\paragraph{DeepOSets-T}
DeepOSets-T integrates a Set Transformer with a DeepONet architecture (Tab.~\ref{tbl:dst-structure}). The Set Transformer is augmented with residual connections and layer normalization, while the DeepONet component incorporates batch normalization. ReLU activations are used throughout the entire model. Following the design of \cite{lee2019set}, the Set Transformer consists of six cascaded Set Attention Blocks (SAB), each with a hidden dimension of 64, followed by attention-based pooling. The output dimension of the Set Transformer is 256. In the DeepONet, both the trunk and branch networks are composed of 8 fully connected layers with a width of 128, and each produces an output of length 500. The variant DeepOSets-TI adopts the inducing points method, replacing all SABs with Induced Set Attention Blocks (ISABs). The total number of parameters in DeepOSets-T is approximately 1.8M, with DeepOSets-TI containing slightly more parameters, depending on the number of inducing points used.

\begin{table}
    \centering
    \caption{DeepOSets-T hyperparameter settings.}
    \label{tbl:model-sizes}
    \begin{tabular}{cc}
        \hline
        \textbf{Set Transformer} & \\
        \# layers&6\\ 
        \# embeding&256\\
        width&64\\
        \# heads&4\\ 
        normalization & layer norm\\
        \hline 
        \textbf{DeepONet} &\\
        \# layers&8\\ 
        width&128\\ 
        \# intermediate&500\\ 
        \hline
        normalization & batch norm\\
        \hline 
        \# params&1.8M\\ 
    \end{tabular}
    \label{tbl:dst-structure}
\end{table}




\bibliography{reference}

\begin{thebibliography}{10}

\bibitem{brown2020language}
T.~Brown, B.~Mann, N.~Ryder, M.~Subbiah, J.~Kaplan, P.~Dhariwal,
  A.~Neelakantan, P.~Shyam, G.~Sastry, A.~Askell, S.~Agarwal, {\em et~al.},
  ``Language models are few-shot learners,'' {\em arXiv preprint
  arXiv:2005.14165}, vol.~1, 2020.

\bibitem{vaswani2017attention}
A.~Vaswani, N.~Shazeer, N.~Parmar, J.~Uszkoreit, L.~Jones, A.~N. Gomez,
  L.~Kaiser, and I.~Polosukhin, ``Attention is all you need,'' {\em arXiv
  preprint arXiv:1706.03762}, vol.~10, p.~S0140525X16001837, 2017.

\bibitem{schmidhuber1987evolutionary}
J.~Schmidhuber, {\em Evolutionary principles in self-referential learning, or
  on learning how to learn: the meta-meta-... hook}.
\newblock PhD thesis, Technische Universit{\"a}t M{\"u}nchen, 1987.

\bibitem{schmidhuber1993neural}
J.~Schmidhuber, ``A neural network that embeds its own meta-levels,'' in {\em
  IEEE International Conference on Neural Networks}, pp.~407--412, IEEE, 1993.

\bibitem{li2017meta}
Z.~Li, F.~Zhou, F.~Chen, and H.~Li, ``Meta-sgd: Learning to learn quickly for
  few-shot learning,'' {\em arXiv preprint arXiv:1707.09835}, 2017.

\bibitem{crawshaw2020multi}
M.~Crawshaw, ``Multi-task learning with deep neural networks: A survey,'' {\em
  arXiv preprint arXiv:2009.09796}, 2020.

\bibitem{javed2019meta}
K.~Javed and M.~White, ``Meta-learning representations for continual
  learning,'' {\em Advances in neural information processing systems}, vol.~32,
  2019.

\bibitem{bommasani2021opportunities}
R.~Bommasani, D.~A. Hudson, E.~Adeli, R.~Altman, S.~Arora, S.~von Arx, M.~S.
  Bernstein, J.~Bohg, A.~Bosselut, E.~Brunskill, {\em et~al.}, ``On the
  opportunities and risks of foundation models,'' {\em arXiv preprint
  arXiv:2108.07258}, 2021.

\bibitem{akyurek2022learning}
E.~Aky{\"u}rek, D.~Schuurmans, J.~Andreas, T.~Ma, and D.~Zhou, ``What learning
  algorithm is in-context learning? investigations with linear models,'' {\em
  arXiv preprint arXiv:2211.15661}, 2022.

\bibitem{garg2023}
S.~Garg, D.~Tsipras, P.~S. Liang, and G.~Valiant, ``What can transformers learn
  in-context? a case study of simple function classes,'' {\em Advances in
  neural information processing systems}, vol.~35, pp.~30583--30598, 2022.

\bibitem{zhang2024trained}
R.~Zhang, S.~Frei, and P.~L. Bartlett, ``Trained transformers learn linear
  models in-context,'' {\em Journal of Machine Learning Research}, vol.~25,
  no.~49, pp.~1--55, 2024.

\bibitem{bai2024transformers}
Y.~Bai, F.~Chen, H.~Wang, C.~Xiong, and S.~Mei, ``Transformers as
  statisticians: Provable in-context learning with in-context algorithm
  selection,'' {\em Advances in neural information processing systems},
  vol.~36, 2024.

\bibitem{xing2024benefits}
Y.~Xing, X.~Lin, N.~Suh, Q.~Song, and G.~Cheng, ``Benefits of transformer:
  In-context learning in linear regression tasks with unstructured data,'' {\em
  arXiv preprint arXiv:2402.00743}, 2024.

\bibitem{liu2024can}
J.~W. Liu, J.~Grogan, O.~M. Dugan, S.~Arora, A.~Rudra, and C.~Re, ``Can
  transformers solve least squares to high precision?,'' in {\em ICML 2024
  Workshop on In-Context Learning}, 2024.

\bibitem{von2023transformers}
J.~Von~Oswald, E.~Niklasson, E.~Randazzo, J.~Sacramento, A.~Mordvintsev,
  A.~Zhmoginov, and M.~Vladymyrov, ``Transformers learn in-context by gradient
  descent,'' in {\em International Conference on Machine Learning},
  pp.~35151--35174, PMLR, 2023.

\bibitem{grazzi2024mamba}
R.~Grazzi, J.~Siems, S.~Schrodi, T.~Brox, and F.~Hutter, ``Is mamba capable of
  in-context learning?,'' {\em arXiv preprint arXiv:2402.03170}, 2024.

\bibitem{tong2024mlps}
W.~L. Tong and C.~Pehlevan, ``Mlps learn in-context on regression and
  classification tasks,'' {\em ICLR 2025}, 2025.

\bibitem{zaheer2017deep}
M.~Zaheer, S.~Kottur, S.~Ravanbakhsh, B.~Poczos, R.~R. Salakhutdinov, and A.~J.
  Smola, ``Deep sets,'' {\em Advances in neural information processing
  systems}, vol.~30, 2017.

\bibitem{lee2019set}
J.~Lee, Y.~Lee, J.~Kim, A.~Kosiorek, S.~Choi, and Y.~W. Teh, ``Set transformer:
  A framework for attention-based permutation-invariant neural networks,'' in
  {\em International conference on machine learning}, pp.~3744--3753, PMLR,
  2019.

\bibitem{li2020neural}
Z.~Li, N.~Kovachki, K.~Azizzadenesheli, B.~Liu, K.~Bhattacharya, A.~Stuart, and
  A.~Anandkumar, ``Neural operator: Graph kernel network for partial
  differential equations,'' {\em arXiv preprint arXiv:2003.03485}, 2020.

\bibitem{lu2021learning}
L.~Lu, P.~Jin, G.~Pang, Z.~Zhang, and G.~E. Karniadakis, ``Learning nonlinear
  operators via deeponet based on the universal approximation theorem of
  operators,'' {\em Nature machine intelligence}, vol.~3, no.~3, pp.~218--229,
  2021.

\bibitem{wang2021learning}
S.~Wang, H.~Wang, and P.~Perdikaris, ``Learning the solution operator of
  parametric partial differential equations with physics-informed deeponets,''
  {\em Science advances}, vol.~7, no.~40, p.~eabi8605, 2021.

\bibitem{wagstaff2019limitations}
E.~Wagstaff, F.~Fuchs, M.~Engelcke, I.~Posner, and M.~A. Osborne, ``On the
  limitations of representing functions on sets,'' in {\em International
  Conference on Machine Learning}, pp.~6487--6494, PMLR, 2019.

\bibitem{keles2023computational}
F.~D. Keles, P.~M. Wijewardena, and C.~Hegde, ``On the computational complexity
  of self-attention,'' in {\em International Conference on Algorithmic Learning
  Theory}, pp.~597--619, PMLR, 2023.

\bibitem{hooper2024kvquant}
C.~Hooper, S.~Kim, H.~Mohammadzadeh, M.~W. Mahoney, Y.~S. Shao, K.~Keutzer, and
  A.~Gholami, ``Kvquant: Towards 10 million context length llm inference with
  kv cache quantization,'' {\em Advances in Neural Information Processing
  Systems}, vol.~37, pp.~1270--1303, 2024.

\bibitem{ge2023model}
S.~Ge, Y.~Zhang, L.~Liu, M.~Zhang, J.~Han, and J.~Gao, ``Model tells you what
  to discard: Adaptive kv cache compression for llms,'' {\em arXiv preprint
  arXiv:2310.01801}, 2023.

\bibitem{jax2018github}
J.~Bradbury, R.~Frostig, P.~Hawkins, M.~J. Johnson, C.~Leary, D.~Maclaurin,
  G.~Necula, A.~Paszke, J.~Vander{P}las, S.~Wanderman-{M}ilne, and Q.~Zhang,
  ``{JAX}: composable transformations of {P}ython+{N}um{P}y programs,'' 2018.

\bibitem{kidger2021equinox}
P.~Kidger and C.~Garcia, ``{E}quinox: neural networks in {JAX} via callable
  {P}y{T}rees and filtered transformations,'' {\em Differentiable Programming
  workshop at Neural Information Processing Systems 2021}, 2021.

\bibitem{paszke2017automatic}
A.~Paszke, S.~Gross, S.~Chintala, G.~Chanan, E.~Yang, Z.~DeVito, Z.~Lin,
  A.~Desmaison, L.~Antiga, and A.~Lerer, ``Automatic differentiation in
  pytorch,'' in {\em NIPS-W}, 2017.

\bibitem{Waskom2021}
M.~L. Waskom, ``seaborn: statistical data visualization,'' {\em Journal of Open
  Source Software}, vol.~6, no.~60, p.~3021, 2021.

\bibitem{Hunter:2007}
J.~D. Hunter, ``Matplotlib: A 2d graphics environment,'' {\em Computing in
  Science \& Engineering}, vol.~9, no.~3, pp.~90--95, 2007.

\bibitem{maron2019provably}
H.~Maron, H.~Ben-Hamu, H.~Serviansky, and Y.~Lipman, ``Provably powerful graph
  networks,'' {\em Advances in neural information processing systems}, vol.~32,
  2019.

\bibitem{klambauer2017self}
G.~Klambauer, T.~Unterthiner, A.~Mayr, and S.~Hochreiter, ``Self-normalizing
  neural networks,'' {\em Advances in neural information processing systems},
  vol.~30, 2017.

\bibitem{kingma2014adam}
D.~P. Kingma, ``Adam: A method for stochastic optimization,'' {\em arXiv
  preprint arXiv:1412.6980}, 2014.

\end{thebibliography}

\end{document}